\documentclass{article}
\usepackage{spconf,amsmath,graphicx,bbding,multirow,url,hyperref}

\usepackage{enumitem}
\setlist{nosep, leftmargin=14pt}

\usepackage{mwe} % to get dummy images

\title{SLAKE: A Semantically-Labeled Knowledge-Enhanced Dataset for Medical Visual Question Answering}
\makeatletter
\newcommand{\printfnsymbol}[1]{%
  \textsuperscript{\@fnsymbol{#1}}%
}
\name{Bo Liu$^{\star}$ 
    \qquad Li-Ming Zhan$^{\star}$\sthanks{Equal contribution.} 
    \qquad Li Xu$^{\star}$\printfnsymbol{1} 
    \qquad Lin Ma $^{\dagger}$ 
    \qquad Yan Yang$^{\ddagger}$ 
    \qquad Xiao-Ming Wu$^{\star}$\sthanks{Corresponding author.}}

\address{$^{\star}$ Department of Computing, The Hong Kong Polytechnic University, Hong Kong \\
    $^{\dagger}$ Department of Ultrasound, West China Hospital of Sichuan University, China\\
    $^{\ddagger}$ Sichuan Academy of Medical Sciences, Sichuan Provincial People's Hospital, China
    }

\begin{document}
%\ninept
%
\maketitle
\begin{abstract}

 Medical visual question answering (Med-VQA) has tremendous potential in healthcare. However, the development of this technology is hindered by the lacking of publicly-available and high-quality labeled datasets for training and evaluation. In this paper, we present a large bilingual dataset, SLAKE, with comprehensive semantic labels annotated by experienced physicians and a new structural medical knowledge base for Med-VQA. Besides, SLAKE includes richer modalities and covers more human body parts than the currently available dataset. We show that SLAKE can be used to facilitate the development and evaluation of Med-VQA systems. The dataset can be downloaded from \url{http://www.med-vqa.com/slake}.
\end{abstract}
\begin{keywords}
Dataset, medical visual question answering, multi-modality fusion. 
\end{keywords}

\section{Introduction} 
\label{sec:intro}
\textbf{}% (AI complete) problem to derive vqa 
% what ismed vqa and its importance
% dataset evolving status and drawbacks
% our dataset and properties
% our dataset is seminal and its excellence

Developing machines that can understand visual content and answer questions like humans is a long-standing goal of AI research. In recent years, visual question answering (VQA) has become an active field of research. Medical visual question answering (Med-VQA) is a domain-specific branch of VQA, where a clinical question comes with a radiology image and the goal is to design a system that can correctly answer the question based on the visual information of the image.

%Developing AI systems that can consciously understand and answer questions about visual information is a long-standing goal of AI research. In recent years, visual question answering (VQA) has been introduced to facilitate the research of this problem. Medical visual question answering (Med-VQA) belongs to a domain-specific branch of VQA, where a clinical question comes with a related radiology image and our purpose is to output the correct answer of this question in natural language based on the visual information of the image.
% (as a form of Visual Turing Test to benchmark performance in computer vision)
% The task simultaneously requires multiple abilities for understanding natural language, visual reasoning and fusing multi-modal features. 
% Compared with general VQA, Med-VQA has a more clear application purpose, helping patients with self-diagnosis, providing clinical decision-making for doctors, especially for inexperienced interns, and teaching everyone who is not familiar with radiology images to master basic knowledge.

\begin{figure}[t]
    \centering
    \includegraphics[width=0.93\linewidth]{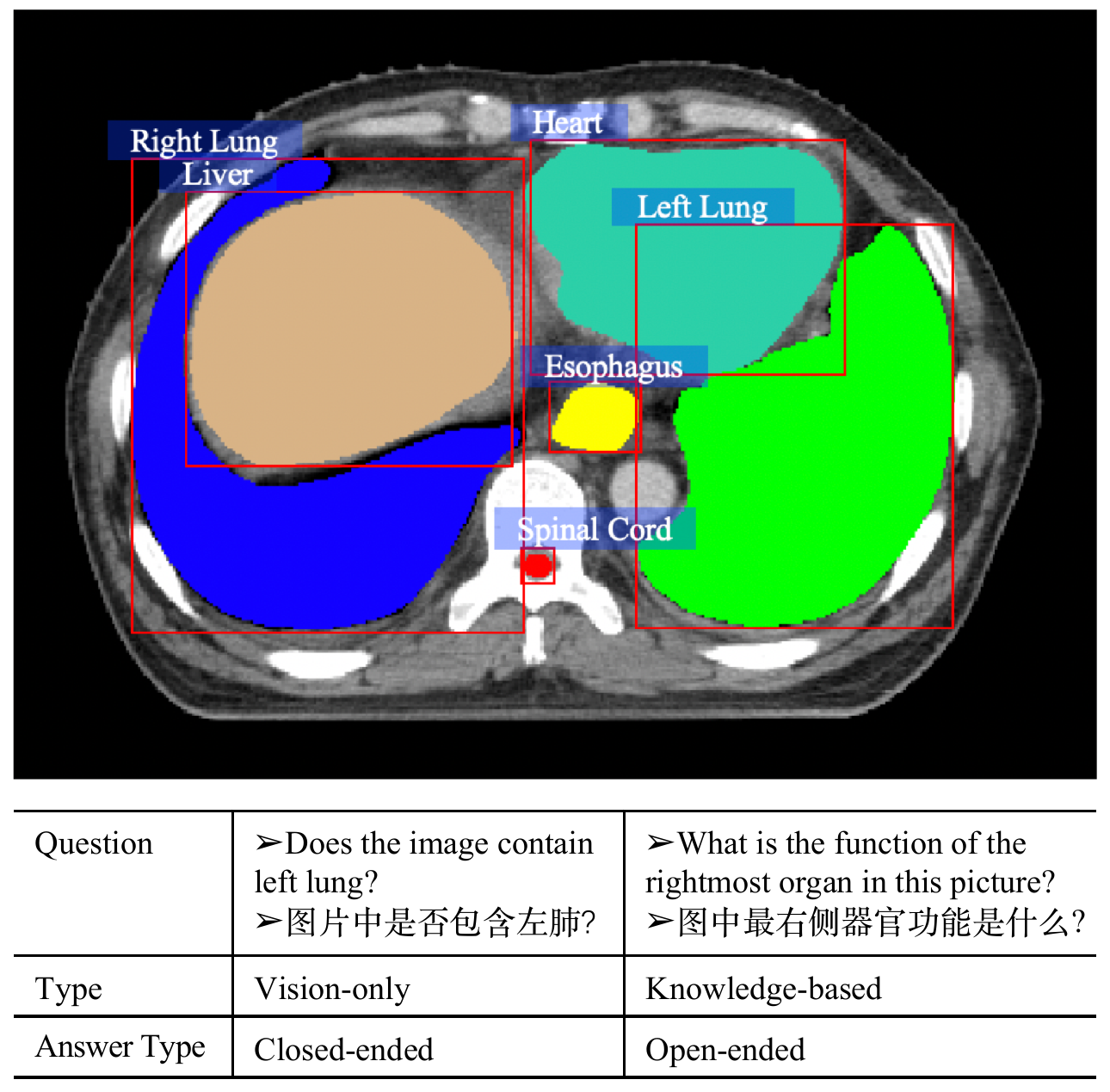}
\caption{Exemplar image and questions of our SLAKE dataset.}

\label{fig:dataset}
\end{figure}

Med-VQA has a wide range of application prospects in healthcare sectors and a broad impact on the wellness of the general public. With a reliable Med-VQA system, patients can easily acquire information about their health and be more engaged in the process of decision making. For doctors, Med-VQA systems can be used to assist diagnosis by providing them a second medical opinion. The systems can also be used in clinical education to train medical professionals. Besides, Med-VQA technology can be potentially integrated into many conversational AI platforms to bring enormous benefits to healthcare industry. 

%Med-VQA has a broad impact on the wellness of the general public. Patients can use Med-VQA systems to acquire preliminary knowledge about their health. Senior doctors can reference the results of Med-VQA systems to assist their diagnosis, and these systems can also be used to the training of junior doctors. Therefore, Med-VQA can help to significantly empower our medical systems. However, the research of Med-VQA is still at its infant stage due to the lack of available datasets. 

% Recently, several datasets for Med-VQA have been proposed and significantly facilitate the development of Med-VQA. 
% Recently, although more and more researchers make contributions to general VQA, which make it popular across the board, due to high barriers of Med-VQA task itself, 
\newcommand{\tabincell}[2]{\begin{tabular}{@{}#1@{}}#2\end{tabular}}
\begin{table*}[ht]
    \centering
    %\small
    \caption{Comparison of SLAKE with VQA-RAD.}
    \label{tab:character}
    \begin{tabular}{cccccc}
    \hline
    \hline
    Dataset & \# Images & \# QA Pairs & Question Type & Language & Knowledge Graph \\
    \hline
    VQA-RAD~\cite{lau2018dataset} & 315 & 3.5K & Vision-only & EN & \XSolidBrush \\
    \hline
    \textbf{SLAKE (Ours)} & 642 & 14K & Knowledge-based \& Vision-only & \tabincell{c}{Bilingual (EN \& ZH) }  &  \Checkmark \\
    \hline
    \end{tabular}
    % \end{threeparttable}
\end{table*}

However, the research of Med-VQA is at an early stage. Unlike VQA in the general domain, where large-scale high-quality datasets~\cite{johnson2017clevr,antol2015vqa} are available, there is a lack of publicly-available and well-annotated datasets for training and evaluating Med-VQA systems. To correctly answer a clinical question about a radiology image, it requires clinical expertise and domain-specific medical knowledge, which makes it difficult to construct a realistic and accurate dataset for Med-VQA. VQA-RAD~\cite{lau2018dataset} is a first step in this direction. To our knowledge, it is the only available dataset with manual annotation, based on which several Med-VQA models have been proposed~\cite{nguyen2019overcoming,ZL-MM-20}. VQA-RAD is a diverse dataset containing a variety of different types of clinical questions, with each question type sufficiently represented. But it does not provide semantic labels, e.g., labeled segmentations of organs and tumors or bounding boxes on objects, which are essential for training a Med-VQA model to find the region of interest in an image to answer complex clinical questions. Moreover, a practical Med-VQA system needs to exploit external knowledge apart from visual content to answer complex compositional questions involving inquires such as ``the functionality of an organ'', ``the cause of a disease'', or ``the treatment of a disease'', which is also not supported in VQA-RAD.

To fill these gaps, we construct a \textbf{s}emantically-\textbf{la}beled \textbf{k}nowledge-\textbf{e}nhanced (SLAKE) dataset with accurate visual and textual annotations and an extendable knowledge base for Med-VQA. It takes our team more than half of a year to complete all the tasks, including building the annotation system, constructing the medical knowledge graph (KG), selecting and labeling images, generating questions, and analyzing the dataset.
% In addition, our dataset contains bilingual questions, i.e., English and Chinese.
% To ensure that the questions and images are more in line with clinical and patient applications, SLAKE is manually constructed by professional physician.
As shown in Figure~\ref{fig:dataset}, for each radiology image, we provide two kinds of visual annotations: \emph{masks} for semantic segmentation and \emph{bounding boxes} for object detection. Besides basic clinical questions, we also design compositional questions that require multiple reasoning steps, and knowledge-based questions like~\cite{wang2018fvqa} that involve external medical knowledge. In general, questions in SLAKE can be categorized as vision-only questions and knowledge-based questions. We provide detailed annotations to distinguish the two types of questions and guide the Med-VQA model to search for answers on the knowledge graph. Besides these new features, SLAKE is designed to be an English-Chinese bilingual dataset to broaden its application range. Further, SLAKE covers more body parts (e.g., neck and pelvic cavity) and more types of questions (e.g., shape and KG-related) than VQA-RAD. A comparison between our SLAKE and VQA-RAD is provided in Table~\ref{tab:character}.

In summary, our contributions are two-fold:
\begin{itemize}
	\item We create SLAKE, a large-scale, semantically annotated, and knowledge-enhanced bilingual dataset for training and testing Med-VQA systems. 
% 	With these abundant information, we expect our dataset can help to build a fertile ground for the development of Med-VQA.
	\item We experiment with representative Med-VQA methods to show that SLAKE can be used as a benchmark to train systems to solve practical and complex tasks. 
\end{itemize} 

% \vspace{-1em}
% In summary, we make three main contributions: (1) the rich visual annotations for each image helping both researchers and models understand medical image reasoning; (2) diverse question types, vision-only, knowledge-based, and bilingual, could make the Med-VQA system meet more requirements and more widely used; (3) wider range of medical images and larger dataset. As a new dataset in Med-VQA, SLAKE fill many blanks for the current domain and provide a solid foundation for development. We hope it can serve as a bridge to a truly widely used medical system in the future. 

% (1) \textbf{rich visual annotations}: for each image in (2) \textbf{diverse question types.} (3) \textbf{extensive medical images.}

\section{The SLAKE Dataset}  
\label{sec:dataset}
% The SLAKE dataset mainly focuses on cross-modal reasoning on the radiology images and medical knowledge graph under the guidance of bilingual question features. 
In this section, we elaborate on the construction of our SLAKE dataset. In general, we ensure the diversity of the dataset in terms of modalities (e.g., CT, MRI, and X-Ray), covered body parts (e.g., head, neck, and chest), and question types (e.g., vision-only, knowledge-based, and bilingual).

\begin{figure}
    \centering
    \includegraphics[width=0.9\linewidth]{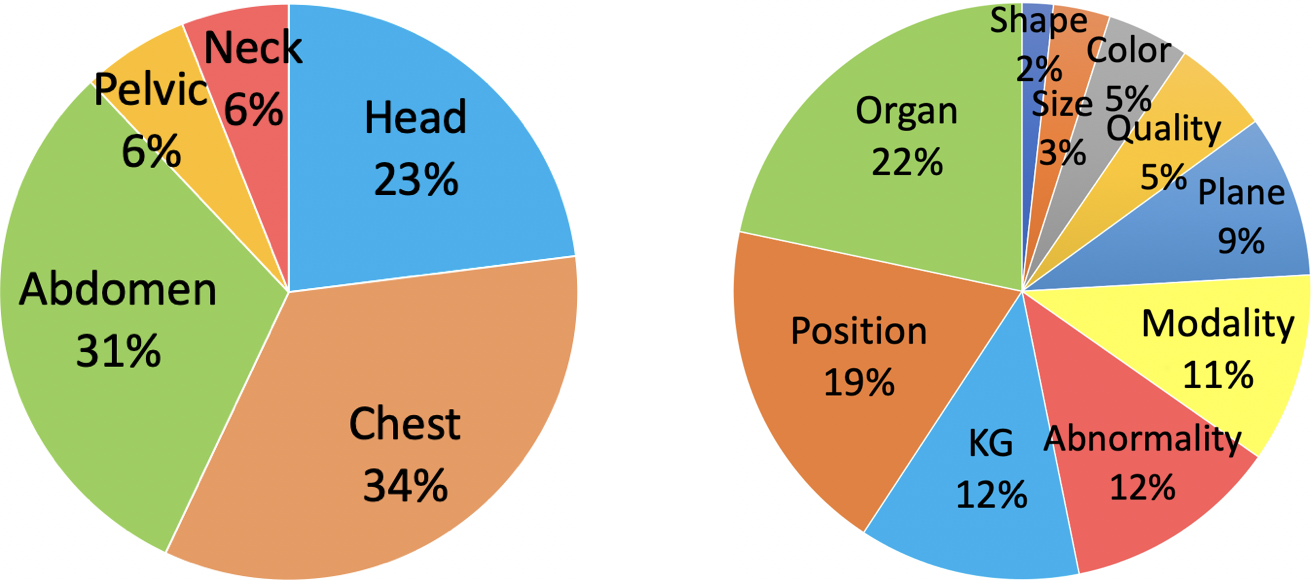}
    \caption{Left: proportions of images of five body parts. Right: distribution of the content types of questions.}
    \label{fig:proportion}
\end{figure}
%reviewer 3
\subsection{Image Acquisition and Annotation} 
\label{sec:selection}
We select radiology images, covering healthy and unhealthy cases, from three open source datasets \cite{simpson2019large}\footnote{\url{http://medicaldecathlon.com}} \cite{wang2017chestx}\footnote{\url{https://nihcc.app.box.com/v/ChestXray-NIHCC}} \cite{CHAOSdata2019}\footnote{\url{https://doi.org/10.5281/zenodo.3431873}}. From \cite{wang2017chestx}, we randomly select $179$ chest X-Ray images and keep the original disease labels. From \cite{simpson2019large} and \cite{CHAOSdata2019}, we randomly choose $463$ single-slice images from $3$\emph{D} volume cases. Then, experienced physicians label organs and diseases as detailed as possible with ITK-SNAP~\cite{py06nimg}\footnote{\url{http://www.itksnap.org}} as shown in Figure~\ref{fig:dataset}. 

% From \cite{wang2017chestx}, we randomly select $179$ chest X-Ray images and keep the original disease labels, such as pulmonary nodules, pneumonia, etc. From \cite{simpson2019large} and \cite{CHAOSdata2019}, we randomly select $463$ images while maintaining the integrity of the data.

% in proportion to the number of organs in five body parts. 

% We , including 140 head CTs or MRIs, 41 neck CTs, 219 chest X-Rays or CTs, 201 abdomen CTs or MRIs, and 41 pelvic cavity CTs. brain edema, brain tumor, esophagus cancer, lung cancer, atelectasis, cardiomegaly, effusion, infiltration, mass, pneumothorax, liver cancer, kidney cancer. 

In total, we annotate 642 images, including 12 diseases and 39 organs of the whole body. The diseases mainly include cancer (e.g., brain, liver, kidney, lung, etc.), and thoracic diseases (e.g., atelectasis, effusion, mass, pneumothorax, etc.). The images include 140 head CTs or MRIs, 41 neck CTs, 219 chest X-Rays or CTs, 201 abdomen CTs or MRIs, and 41 pelvic cavity CTs. The distribution is shown in Figure~\ref{fig:proportion} (Left). Among these images, there are 282 CTs, 181 MRIs, and 179 X-Rays. All CTs and MRIs are axial single-slice. The number of images for each body part is set based on the complexity of the body part. For example, the number of diseases and organs in abdomen is much more than that in neck, so there are more images of abdomen than neck in the dataset.

\subsection{Knowledge Graph Construction}
%traverse
To answer questions that require external medical knowledge, we construct a medical knowledge graph centered on organs and related diseases, which are the main objects of radiology images. 
We extract a set of 52.6K triplets \emph{\textless head, relation, tail\textgreater} with medical knowledge from OwnThink\footnote{\url{https://www.ownthink.com}}, a large-scale knowledge base built on Wikipedia. Here, \emph{head} and \emph{tail} are entities such as organ, disease, etc., and \emph{relation} represents the relationship between entities, such as function or treatment. Then, we traverse the set to retrieve triplets related to organs and the corresponding diseases. We further clean the data by manually filtering out some entities that are not presented in medical images such as gastritis and nephritis.
% However, many of them are not useful because the entities could not be seen on radiology images, such as gastritis, nephritis. Therefore, we filter the triplets to make them more suitable for knowledge-based visual reasoning.

\begin{table}[t]
    \centering
    \small
     \caption{Statistics of questions in our SLAKE dataset.}
      \begin{tabular}{c|cccc}
  \hline
  \hline
  &\textbf{Training set}&\textbf{Validation set}&\textbf{Test set}\\
  \hline
   
     Plane       &	931  &   173  &	176	\\
     Quality    &	535  &	109  &	118	\\
     Modality    &	1072  &	203  &	217   \\
     Position	 &  1876	&   412 &	390  \\
     Organ      &	2125 &	462	&   454	\\
     KG          &	1202 &	278	&   260 \\
     Abnormal	 &  1230	&   245  &	221   \\
     Color      &	424	&   108	&   115   \\
     Shape       &	157	&   42	&   46   \\
     Size      &	297	&   77 &	73   \\
    \hline
    Total&{\bf 9849}&{\bf 2109}&{\bf 2070}\\
    \hline
    \hline
    \end{tabular}
    \label{tab:statistic}
\end{table}

Next, in order to extensively cover frequently referenced knowledge, we refine the filtered triplets with the following rules: (1) The triplets about an organ must describe its function or body system; (2) The triplets about a disease must describe the symptoms, locations, causes, treatment or prevention methodologies. Some examples are shown in Table~\ref{tab:examle_kg}.

%(1) For each organ, the triplet set must contain the organ's function and body system; (2) For the diseases, the triplet sets must contain symptoms, locations, causes, treatment and prevention methodologies, as shown in Table~\ref{tab:examle_kg}. 

Finally, we make the triplets bilingual and obtain $2603$ triplets in English and $2629$ triplets in Chinese.

\subsection{Question Generation}
\label{sec:qg}

%reviewer2 
Questions are proposed by experienced doctors. To accelerate this process, we develop an annotation system. In this system, we first pre-define a question template for each body part (i.e., head, neck, chest, abdomen, and pelvic cavity). Then, we define ten different content types (e.g., \emph{modality}, \emph{position}, \emph{color}) for the questions, as shown in Table~\ref{tab:statistic} and Figure~\ref{fig:proportion} (Right).
% more clear 
In each template, we provide many candidate questions for each content type. For example, the candidate question for a head image with the content type \emph{organ} may be ``Is this a study of the head?'' or ``What organ system is imaged?''. Physicians could choose those candidate questions or amend or even rewrite them entirely based on their personal clinical experience. The flexibility of our annotation system ensures the question diversity of SLAKE. Note that because we provide different candidates for bilingual questions, the number and content of them in our dataset are not the same. 

%Additionally, because the candidate questions are randomly selected from templates, the number of questions varies from language to language and there is no translation relationship between bilingual questions of one image.

% change <vhead>
Moreover, we provide semantic label for each question. Specifically, we use  \emph{\textless vhead, \_ , \_\textgreater} (\emph{vhead} is a placeholder) to denote vision-only questions. For a knowledge-based question like ``Which organs in this image belong to the digestive system?'', we denote it as \emph{\textless vhead, belong to, digestive system\textgreater}. Such labeling helps to distinguish question type and identify the part of the question involving external knowledge.

Besides, recent studies~\cite{agrawal2016analyzing,zhang2016yin} have shown that VQA models may be susceptible to the statistical bias of answer distribution of the datasets. To mitigate the inherent bias of SLAKE, we make the answers balanced in general such that the VQA model will not be biased to the most popular answer in the dataset.
For example, for the question ``Is this a study of the abdomen?'',  we make sure this question is asked with abdomen images and non-abdomen images with $50-50$ chance, thereby keeping the numbers of ``Yes'' and ``No'' balanced.
% In total, the number of answers ``Yes'' and ``No'' for it would be balanced as long as the number of images in the abdomen and other parts is proportional. be trapped by the low hanging fruit of merely outputting

\begin{table}[t]
 
  \centering
  \small
%   \fontsize{7}{8}\selectfont
%   \begin{threeparttable}
  \caption{Examples of our medical knowledge graph.}
  \label{tab:examle_kg}
   \centering
  \small
  \begin{tabular}{c|c}
  \hline
    \hline
     & Examples \\  \hline
    \multirow{3}{*}{Organ} &  {\emph{\textless Heart, Function, Promote blood flow\textgreater}}\\ 
     &  {\emph{\textless Kidney, Belong to, Urinary System\textgreater}}\\
     &  {\emph{\textless Duodenum, Length, 20-25cm\textgreater}}\\
    \hline
    \multirow{5}{*}{Disease} &  {\emph{\textless Pneumonia, Location, Lung\textgreater}}\\ 
     &  {\emph{\textless Lung Cancer, Cause, Smoke\textgreater}}\\ 
     &  {\emph{\textless Brain Tumor, Symptom, Visual impairment\textgreater}}\\ 
     &  {\emph{\textless Cardiomegaly, Treatment, Medication\textgreater}}\\ 
     &  {\emph{\textless Atelectasis, Prevention, Exercise\textgreater}}\\ 
    \hline
    \hline
    \end{tabular}
    % \end{threeparttable}
\end{table}

\subsection{Dataset Splitting}
Here, we describe how to divide the obtained 642 images with 14,028 question-answer pairs and 5232 medical knowledge triplets for the training and evaluation of Med-VQA models. 

In general, the splitting aims to provide a reliable measure of the generalization ability of the model trained on our dataset.
Specifically, we split the dataset into training ($70\%$), validation ($15\%$), and test ($15\%$) sets at the image level. The images in our dataset are split with the 75:15:15 ratio in each of the $8$ categories: ``head CT'', ``head MRI'', ``neck CT'', and ``chest X-Ray'', ``chest CT'', ``abdomen CT'', ``abdomen MRI'', and ``pelvic cavity CT''. Note that we only divide the images but the questions associated with each image are not split. 

%reviewer 3, split

% To reasonably examine the performance of a model on SLAKE, we split the dataset into training ($70\%$), validation ($15\%$), and test ($15\%$) sets at the image level. 
% In general, we split the dataset into training ($70\%$), validation ($15\%$), and test ($15\%$) sets at the image level. 
% % change
% Since the images are about different body parts and in different modalities, randomly splitting them may result that there are some images in the test set whose features are never seen in the training set~\cite{teney2016zero}. It would then be difficult, if not possible, for the model to correctly answer the accompanied questions during testing. Thus, we split the images within each of the $8$ categories ``head CT'', ``head MRI'', ``neck CT'', ``chest X-Ray'', ``chest CT'', ``abdomen CT'', ``abdomen MRI'', and ``pelvic cavity CT'', with the 75:15:15 ratio. 
% To ensure the test set can be representative enough for measuring the generalization , 

% In our dataset, we have $7$ kinds of units and independently sample images from them according to the above proportion. 
Besides, since VQA is usually formulated as a classification task~\cite{nguyen2019overcoming,ZL-MM-20,yang2016stacked},
%usually formalized as classification problem we fooloew the convention
we follow the convention and make sure answers in the test set must appear in the training set. Finally, the images are split into $450$ for training, $96$ for validating, and $96$ for testing. The number of questions of different type in each set is shown in Table~\ref{tab:statistic}. 

% To avoid inherent bias in the dataset, we keep the number of specific counterpart answers (like ``YES'' or ``NO'') balanced in general such that statistical models will not be trapped by the low hanging fruit of merely outputting the answer that occupies the larger proportion in the dataset

\section{Experiments}   
\label{sec:baseline}
% \begin{Fig.}[t]
%     \centering
%     \includegraphics[width=0.92\linewidth]{structure.png}
% \caption{The structure of system designed for SLAKE.}
% \label{fig:structure}
% \end{Fig.}

% where to clarify the meaning of the open-ended and closed-ended

% intro baseline -> huge gap for using -> questions balanced -> annotations useful
\begin{figure}
    \centering
    \includegraphics[width=0.95\linewidth]{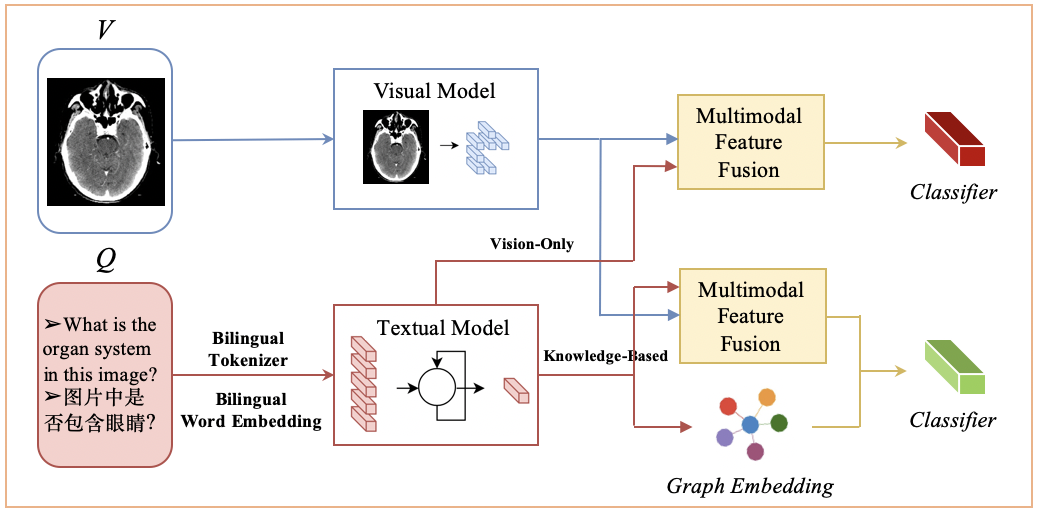}
    \caption{The Med-VQA framework on our SLAKE dataset.}
    \label{fig:structure}
\end{figure}

% In this section, we conduct assorted experiments to comprehensively evaluate our SLAKE dataset proposed in Section~\ref{sec:dataset}. In order to correctly handle different types of tasks in SLAKE, we design a structure that simultaneously processes vision-only and knowledge-based bilingual questions. Then we apply current state-of-the-art methods to obtain reliable results as strong baselines, revealing the challenge of dataset. Meanwhile, it can be seen by analyzing the accuracy of questions with different answer types that the number of questions is balanced. Finally, we do the experiments to verify that the semantic annotations for both vision and text we provide are effective. 
In this section, we conduct extensive experiments to comprehensively evaluate our SLAKE dataset. To be elaborated later, Table~\ref{tab:exp} demonstrates the usefulness and the challenge of SLAKE with commonly used Med-VQA methods. To show the effectiveness of the constructed medical knowledge graph, we conduct an ablation study presented in Table~\ref{tab:KG_exp}.
\subsection{Experiment Setup}
\label{sec:setup}
The pipeline of our experiments is illustrated in Figure~\ref{fig:structure}. We experiment with a commonly used Med-VQA framework, stacked attention network (SAN)~\cite{yang2016stacked}, on SLAKE.
% Considering modules selection in whole structure, 
% In general, the architecture of our experiments is shown in Figure~\ref{sec:baseline}
We use VGG16~\cite{simonyan2014very} to extract visual features from radiology images. For bilingual questions, we first design a bilingual tokenizer to create bilingual word embeddings for the English questions and Chinese questions respectively. 
Then, a 1024\emph{D} - LSTM is applied to extract textual semantics from these embeddings and classify types of questions. There are two sub pipelines in Figure~\ref{fig:structure}. Given the extracted visual and textual features, vision-only tasks will be directed to the multimodal fusion module of SAN to create fused features for classification. For knowledge-based tasks, question-related embeddings extracted from the knowledge graph will be combined with the multimodal fused features for classification. 
% The experiments are executed on an Ubuntu server with Titan XP GPUs. 
% When meet the knowledge-based ones, we would combine additional question related embedding extracted from knowledge graph. 

% Finally, the MLPs could be adopted as classifier to reason final answers.In this paper, the metric we used for evaluating dataset is accuracy.

\subsection{Dataset Analysis}
We report the results for vision-only and knowledge-based questions in Table~\ref{tab:exp} and Table~\ref{tab:KG_exp} respectively. Answers of ``closed-ended'' questions are limited multiple-choice options,  while answers of ``open-ended'' questions are free-form texts. Open-ended questions are generally harder to answer than closed-ended ones. 
%Open-ended questions are generally harder to answer than closed-ended ones. 
\begin{table}[t]
    \centering
    \caption{Accuracy for vision-only questions (\%).}
    \label{tab:exp}
    \begin{tabular}{cc|ccc}
    \hline
    \hline
    \multirow{2}{*}{Language} & \multirow{2}{*}{Models}  & \multirow{2}{*}{Overall} &  \multirow{1}{*}{Open-}  & \multirow{1}{*}{Closed-} \\
          &     &    &   ended &  ended   \\
    \hline
    % &ResNet+SAN &  73.85    &     73.48      &     74.39    \\
    % ResNet+BAN &  74.97    &     74.24      &     \textbf{76.01}         \\
    \multirow{2}{*}{English} & VGG+SAN&   72.73    &     70.34      &     76.13\\
        &   VGG\emph{seg}+SAN &   \textbf{75.36}    & \textbf{72.20}&    \textbf{79.84} \\
    \hline
    Chinese & VGG+SAN& 74.27 & 73.64& 75.20 \\
    \hline
    % \multirow{2}{*}{Knowledge-Based}&\multirow{2}{*}{English} & VGG+SAN & 70.65 & 73.65& 63.95\\
    % & & VGG+SAN+KG & 73.09 & 71.65 & 76.32\\
    \hline
    \end{tabular}
\end{table}
\begin{table}[t]
    \centering
    \caption{Accuracy for knowledge-based questions (\%).}
    \label{tab:KG_exp}
    \begin{tabular}{cc|c}
    \hline
    \hline
     Language & Models    &   Overall   \\
    \hline
    \multirow{2}{*}{English}& VGG+SAN & 70.27 \\ %& 73.65& 63.95\\
        &VGG+SAN+KG & \textbf{72.30} \\% & 71.65 & 76.32\\
    \hline
    Chinese & VGG+SAN+KG & 75.01 \\ %& 73.65& 63.95\\{74.91} \\% & 71.65 & 76.32\\
    \hline
    \hline
    \end{tabular}
\end{table}
%huge gap jieshi

\textbf{Vision-only questions.} In Table~\ref{tab:exp}, we report the results in accuracy for vision-only questions in both English and Chinese. Compared with VQA in the general domain, clinical questions in Med-VQA need to be answered as accurate as possible because they relate to health and safety. It can be seen that the baseline models achieve accuracy of around $73\%$ which is still far away from practical use in the medical domain. There is a wide gap between this and clinical standard, which shows that SLAKE is challenging. Moreover, it can be seen that the overall accuracy is roughly the average of those of open-ended and closed-ended questions, proving that the question distribution of SLAKE is balanced.
        
Besides, to demonstrate the usefulness of the semantic visual annotations elaborated in Section~\ref{sec:selection}, we design another model, VGG\emph{seg}+SAN. First, we pretrain a fully convolutional network (FCN) with VGG backbone by the segmentation task of radiology images with respect to the \emph{mask} labels in the training set.
Then, we initialize the VGG backbone in the Med-VQA model with the pretrained parameters. The overall accuracy increases from $72.73\%$ to $75.36\%$ with a $2.6\%$ improvement, which shows that our semantic visual annotations could improve the reasoning abilities of the model.

\textbf{Knowledge-based questions.} We leverage the self-built medical knowledge graph to answer knowledge-based questions. First, we randomly initialize an embedding for each entity in the knowledge graph and use the TransE~\cite{bordes2013translating} method to enforce the embeddings of the entities in each triplet, \emph{\textless head, relation, tail\textgreater}, to satisfy: $ \emph{head} \boldsymbol{+} \emph{relation} \boldsymbol{\approx} \emph{tail}$. Then, based on the semantic textual annotations (Section~\ref{sec:qg}), we train two LSTMs to predict the words for the ``relation'' and ``tail'' of a question separately. Next, we find the corresponding entity embeddings of the relation and tail from the graph and use them to obtain the head entity embedding based on the above approximate equation, which is then combined with the fused multimodal features for final prediction. The result is reported in Table~\ref{tab:KG_exp}.
For comparison, we also try to predict answers without using the knowledge graph. The result is $2.0\%$ lower, indicating that the constructed knowledge graph is informative and it is helpful to leverage external structural knowledge to tackle knowledge-based questions.

% which learns a embedding for knowledge graph.
% For each triplets \emph{\textless head, relation, tail\textgreater} in the knowledge graph, transE learns the embedding $h$, $r$ and $t$ for head, relation and tail respectively, which maintain $h + r = t$. % implementation of thunlp
% We use Stacked Attention Network(SAN) \cite{kim2018bilinear} to extract multimodal features of images and questions. Here we use VGG16 to extract image feature and LSTM for question feature.
% In the meanwhile, we train 2 LSTMs to classify the relation and the tail in questions separately so that we can combine the embedding of knowledge graph with the multimodal feature.

% The procedure is shown in bottom-right part of Fig.  \ref{fig:structure}. The multimodal feature and the embedding are fed into a MLP and the model can give the answer to a question. For comparison, we use a single SAN as baseline. we run two model 5 times seperately and report the average accuracy.
% The result is shown in Table \ref{tab:KVQA_res}. 

\section{Conclusion} 
\label{sec:conclusion}
% In this paper, we introduced XMVQA for bilingual reasoning on radiology images and knowledge graph. We described its construction process and composition in detail. For each task in dataset, we did experiments and provided strong baselines. As a new dataset in Med-VQA, it filled many gaps for this domain. We strongly hope that XMVQA could promote the development of medical visual question answering.

We have introduced SLAKE, a new large bilingual dataset to facilitate the training and evaluation of Med-VQA systems.
% and to advance the research of Med-VQA fundamentally. 
SLAKE is a diverse and balanced dataset containing rich visual and textual annotations and a unique medical knowledge graph, which allows the development of more powerful Med-VQA systems. Remarkably, our experiments show that the semantic annotations and external knowledge can significantly improve the performance of standard Med-VQA models. We hope SLAKE will serve as a stepping stone to push forward the research of Med-VQA.
% To avoid inherent bias in the dataset, we keep the number of specific counterpart answers (like ``YES'' or ``NO'') balanced in general such that statistical models will not be trapped by the low hanging fruit of merely outputting the answer that occupies the larger proportion in the dataset.
% We have struck the diversity of SLAKE in terms of modalities (e.g., CT, MRI and X-Ray), and body parts (e.g., head, neck and chest), and question types (e.g., vision only, knowledge-based, and bilingual).  and our dataset will be released after the acceptance

\clearpage
\vfill
\pagebreak

\section{Compliance with Ethical Standards} 
\label{sec:ethical}
This research study was conducted retrospectively using human subject data made available in open access by~\cite{simpson2019large,CHAOSdata2019,wang2017chestx}. Ethical approval was not required as confirmed by the license attached with the open access data.

\section{Acknowledgments} 
\label{sec:acknowledgments}
We would like to thank the anonymous reviewers for their helpful comments. Thanks to Lau et al~\cite{lau2018dataset} for their pioneering work in Med-VQA, NIH Clinical Center for sharing their open access dataset~\cite{wang2017chestx}, and all the doctors and medical students who helped with this research. This research was supported by the grant of P0030935 (ZVPY) funded by PolyU (UGC).

\bibliographystyle{IEEEbib}
\bibliography{refs}
\end{document}